\documentclass[runningheads]{llncs}
\usepackage{graphicx}
\usepackage{float}
\begin{document}
\title{{Differentiable Reasoning over Long Stories} \\
{\large Assessing Systematic Generalisation in Neural Models}\\}
%
\author{Wanshui Li\inst{1}\orcidID{0000-0002-0709-1860} \and
Pasquale Minervini\inst{1}\orcidID{0000-0002-8442-602X}}
\institute{University College London, London WC1E 6BT, UK}
\maketitle              
\begin{abstract}
Contemporary neural networks have achieved a series of developments and successes in many aspects; however, when exposed to data outside the training distribution, they may fail to predict correct answers. In this work, we were concerned about this generalisation issue and thus analysed a broad set of models systematically and robustly over long stories. Related experiments were conducted based on the CLUTRR, which is a diagnostic benchmark suite that can analyse generalisation of natural language understanding (NLU) systems by training over small story graphs and testing on larger ones. In order to handle the multi-relational story graph, we consider two classes of neural models: "E-GNN", the graph-based models that can process graph-structured data and consider the edge attributes simultaneously; and “L-Graph”, the sequence-based models which can process linearized version of the graphs. We performed an extensive empirical evaluation, and we found that the modified recurrent neural network yield surprisingly accurate results across every systematic generalisation tasks which outperform the modified graph neural network, while the latter produced more robust models.

\end{abstract}
\section{Introduction}

Since the earliest “artificial neuron” model proposed based on studies of the human brain and nervous system, neural networks have achieved a series of developments and even notable successes in some commercially essential areas, such as image captioning, machine translation and video games. However, several shortcomings in modern deep neural networks are emphasised by Garnelo and Shanahan \cite{garnelo2019reconciling}:
\begin{itemize}
  \item Data inefficiency and high sample complexity. In order to be effective, contemporary neural models usually need large volumes of training data, such as BERT~\cite{Devlin_Chang_Lee_Toutanova_2018} and GPT2~\cite{radford2019language} with the pre-trained basis which comes from an incredible amount of data.
  \item Poor generalisation. It is a challenge to predict correct answers when neural models are evaluated on examples outside of the training distribution, and even just small invisible changes to the inputs can entirely derail predictions~\cite{goodfellow2014explaining}.
\end{itemize}
Meanwhile, the ability of NLU systems is increasingly being questioned --– and neural networks more generally --– to generalise systematically and robustly \cite{bahdanau2018systematic}.  For instance, the brittleness of NLU systems to adversarial examples \cite{DBLP:journals/corr/JiaL17}, the failure of exhibiting reasoning and generalisation capabilities but only exploiting statistical artefacts in datasets \cite{Gururangan_Swayamdipta_Levy_Schwartz_Bowman_Smith_2018}, the difficulty of incorporating the statistics of the natural language instead of reasoning on those large pre-trained models \cite{DBLP:journals/corr/abs-1802-05365} and the substantial performance gap of generalisation and robustness between state-of-art NLU models and a Graph Neural Networks (GNN) model \cite{sinha2019clutrr}. It seems that modern neural models capture the wrong pattern and do not understand the content of data, which is far away from the expected reasoning ability mimicking from the human brain.  In other words, both of this aspect should be emphasised so that we can build up a better model.

In order to evaluate and compare each model's ability of systematic generalisation and robust reasoning, we use the benchmark suite name Compositional Language Understanding and Text-based Relational Reasoning (CLUTRR), which contains a broad set of semi-synthetic stories involving hypothetical families, and the goal is to reason the relationship between two entities when given a story \cite{sinha2019clutrr}. Actually, \cite{sinha2019clutrr} already analysed several models, including those shocking baselines models we have known such as BERT and Graph Attention Networks (GAT) \cite{velivckovic2017graph}. They found that existing NLU systems show poor results both on generalisation and robustness comparing GAT who directly works on symbolic inputs. For instance, interestingly line graphs in the paper showed us that in most situations, BERT stays the lowest accuracy among all text-based models and also a graph neural model, GAT; when trained on noisy data, only GAT can effectively capture the robust reasoning strategies. Both of these phenomena show us that there is a gap between unstructured text inputs and structured symbolic inputs.

Therefore, motivated by that unexpected results and inspired by the neuro-symbolic reasoning models \cite{garcez2015neural,evans2018learning,garnelo2019reconciling}, we explore two types of models, the graph-based model and the sequence-based model. Each of them has different forms of symbolic inputs, and they are trained over the CLUTRR datasets to evaluate their generalisation and robustness performance compared with texted-based models. Briefly speaking, graph-based models are those models with a graph as input, such that in CLUTRR datasets, entities and relationships are modelled as nodes and edges to form graphs. To deal with graph-type input, we usually use GNN models such as GAT, Graph Convolutional Networks (GCN) \cite{kipf2016semi}. In fact, their architectures need to be modified so that we can feed the multi-relational graph into models. For sequence-based models, they are the revised version of those typical sequence encoding models such as Convolutional Neural Networks (CNN) \cite{LeCun1989}, Recurrent Neural Networks (RNN) \cite{Rumelhart_Hinton_Williams_1986}. Their entities and relationships are modelled into graphs and then concatenated into sequences.

\section{Related Work} \textbf{Neuro-symbolic Models.} Recent advances in deep learning show an improvement of expressive capacity, and significant results have been achieved in some perception tasks, such as action recognition and event detection. Nevertheless, it is widely accepted that an intelligent system needs combining both perception and cognition. Highly cognitive tasks, such as abstracting, reasoning and explaining, are closely related to symbol systems which usually cannot adapt to complex high-dimensional space. Neuro-symbolic models combine the advantages of the deep learning model with symbolic methods, thereby significantly reducing the search space of symbolic methods such as program synthesis \cite[cf.]{gan2017vqs,yi2018neural}.

There are many applications and methods of neuro-symbolic reasoning in natural language processing (NLP), basically involved complex question answering (QA), that is, given a complicated question, the answer is inferred from the context. Usually, the process is to split the tricky question into several sub-questions and then to use the neuro-symbolic reasoning model to get the results, which requires the abilities of question understanding, information extraction of context, and symbolic reasoning.

For instance, Gupta et al.\cite{gupta2019neural} proposed Neural Module Networks (NMN) to solve tricky question answering tasks, where the reasoning process over texts involves natural language reasoning and symbolic reasoning. Natural language reasoning can be seen as the process of information extraction from texts, and symbolic reasoning is the reasoning process based on the extracted, structured information. Compared with NMN customising modules based on tasks, Compositional Attention Networks (MAC) \cite{Hudson_Manning_2018} is a soft-attention architecture (i.e. computing all corresponding attention weight with all the data) providing a more universal and reusable architectures with shared parameters, and it is an end to end differentiable. Neural State Machine (NSM) \cite{hudson2019learning} has a similar reasoning mechanism that in MAC, but its expression way is the probability distribution of a scene graph based on the given content, which shows powerful generalisation capacity under multi-tasks. Neuro-Symbolic Concept Learner (NS-CL) \cite{mao2019neuro} has three modules: extract and express target in fixed-length vectors, similarity comparison for the object and contents, execution and analysis with a curriculum learning approach, where the modules are efficient and even achieve good results over a fewer quantity of data.

By combining neural networks with symbolic systems, models can be improved effectively in their shortcomings aspects, such as promoting data efficiency due to reusable property in multiple tasks; becoming interpretable or human understanding to some extent because of human-like thinking process; facilitating generalisation capacity because of high-level, abstract representations. In our work, we modify the architectures of GNN and sequence encoding models (i.e. CNN, RNN) to achieve the neuro-symbolic reasoning process (see Chapter \ref{char3}), and based on their types of input, we separate them into two classes: the graph-based model and the sequence-based model.

\section{Assessing Systematic Generalisation \label{char3}}

Two types of model, the E-GNN (graph-based) model and the L-Graph (sequence-based) model, are used to work on CLUTTR datasets to evaluate the performance of systematic generalisation and robust reasoning as well.

\subsection{Datasets} As in \cite{sinha2019clutrr}, we use the same pre-generated datasets \footnote{https://github.com/facebookresearch/clutrr/} to evaluate our models and it is also easier and more obvious to compare our models' performance with that in the paper. There are totally six datasets separated into two groups: two of them (named "data\_089907f8" and "data\_db9b8f04") are used for systematic generalisation evaluation, and the rest of them (named "data\_7c5b0e70", "data\_06b8f2a1", "data\_523348e6" and "data\_d83ecc3e") are used for robust reasoning evaluation.

\begin{figure}[H]
\includegraphics[width=1\textwidth]{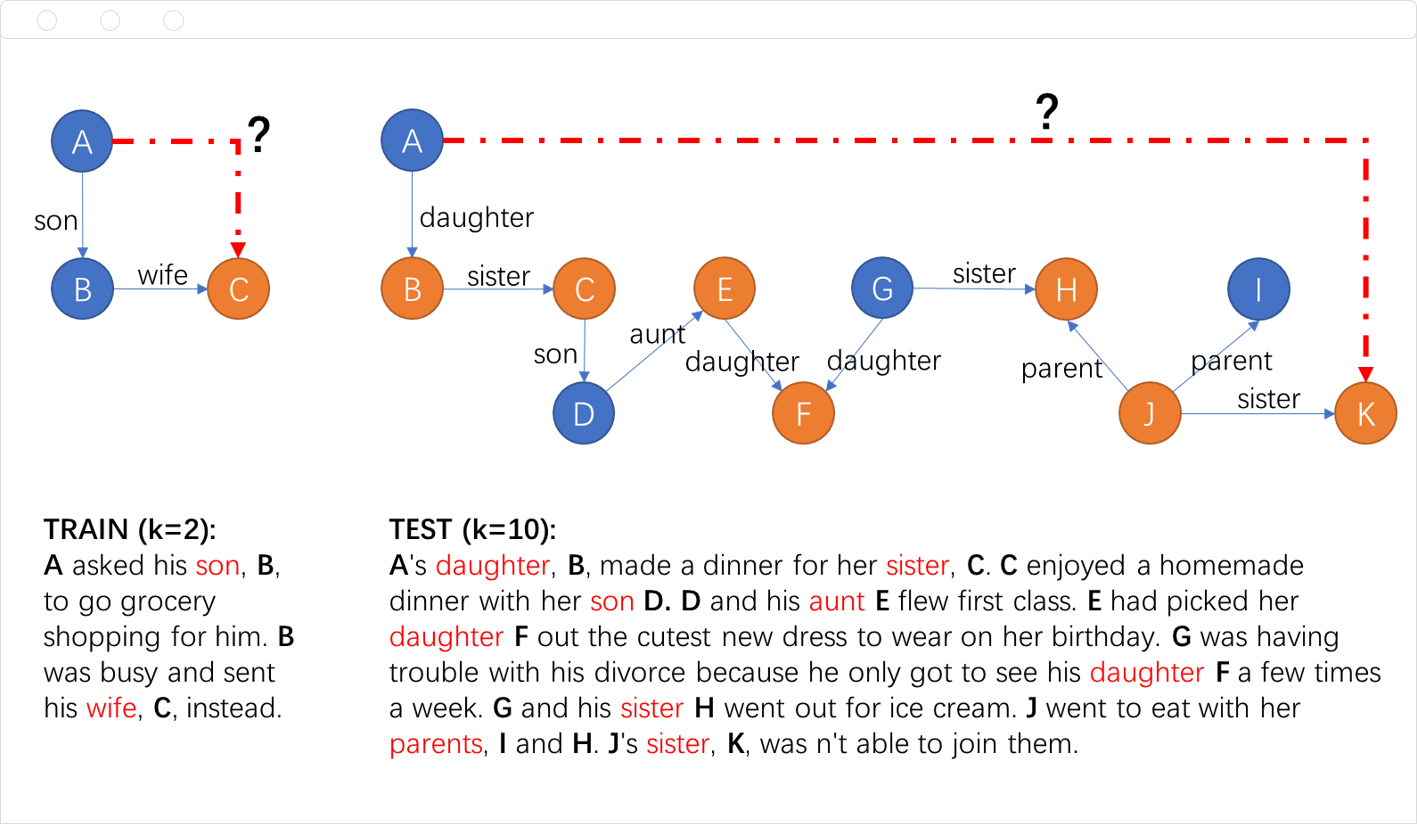}
\caption{An example of a train and test instance in CLUTRR. Training over instance with clauses of length k=2 (left), and testing over instance with clauses of length k=10 (right). All names are replaced with capital alphabet since replaceable and unrelated in the process; relationships are shown between two nodes beside the line. Task (or query) is to identify the relationship between two nodes linked by red dash based on the given supporting facts.}
\label{Fig.ttex} 
\end{figure}

We test our models' systematic generalisation capacity with clauses of length $k$ = 2, 3 $\cdots$, 10 for both datasets but training process has two regimes: clauses of length in "data\_089907f8" are $k$ = 2, 3 and in "data\_db9b8f04" are $k$ = 2, 3, 4. An example of a train and test instance in CLUTRR in \textbf{Fig. \ref{Fig.ttex}}.

\subsection{E-GNN \label{Graph-based}} In order to handle multi-relational graphs and better learn the relationship between two entities, we modified the normal Graph Neural Networks (GNN) architecture. Entities and relationships are modelled as nodes and edges, respectively in the graph, and we concatenate relationships information on the entities so that the model can also consider the edge attributes during encoding. In other words, the graph-based mechanism is based on the supporting facts to embed a query and seek the target, so we can also present this type of model in the form of:
$$\hat{y} = softmax\left(\mathbf{W}[emb(\mathcal{F}) \, \Vert \, emb(\mathcal{Q}\, \vert \, \mathcal{F})]\right)$$, where $emb(\cdot)$ denotes an embedding process working on a set of ground facts, such that the supporting facts ($\mathcal{F}$) or a query conditioning on supporting facts ($\mathcal{Q\, \vert \,F}$); $"\, \Vert \,"$ means a concatenation of two items; $\mathbf{W}$ represents weight matrix; and $\hat{y}$ represents the distribution of all possible relationships we get after softmax function, and then we will choose the highest probability relationship type as the predicted target.

After processing with GNN models, $emb(\mathcal{F})$ is the representation of each node in the graph with a shaped [$B \, \times \, N \, \times \, emb_{dim}$], where $B$ is the batch size, $N$ is the number of nodes and $emb_{dim}$ is the embedding dimension. $emb(\mathcal{Q\, \vert \,F})$ is the representation of each query and it comes from $emb(\mathcal{F})$. Here, we gather the representation of needed nodes in the query from $emb(\mathcal{F})$ and it shaped in [$B \, \times \, (2*emb_{dim})$], where "2" means two entities/nodes in a query. Besides, when we conduct concatenating, $emb(\mathcal{F})$ is reshaped into [$B \, \times \, emb_{dim}$] by taking average on the nodes, and we get the shape [$B \, \times \, (3*emb_{dim})$] in the end. To discuss this in more detail, we need to look at the message passing framework used in the GNN model. 

\subsubsection{Message Passing Framework} When we handle the graph data, we normally will conduct convolution operation in a neighbourhood aggregation or message passing scheme, which can be expressed as

$$\mathbf{x}_i^{(k)} = \textbf{UPDATE} \left(\mathbf{x}_i, \textbf{ AGGR}_{j\in \mathcal{N}(i)} \textbf{ MESSAGE}^{(k)} (\mathbf{x}_i^{(k-1)}, \mathbf{x}_j^{(k-1)}, \mathbf{e}_{ij}) \right) $$

, where $\mathbf{x}_i^{(k-1)}$ denotes the state of the current node $i$ in the $(k-1)^{th}$ layer; $\mathbf{x}_j^{(k-1)}$ denotes the state of the neighbour node $j$ in the $(k-1)^{th}$ layer; $\mathbf{e}_{ij}$ denotes the edge features from node $j$ to node $i$; $\mathcal{N}(i)$ denotes the neighbourhood set of node $i$ (i.e. 1-hop neighbours); $\textbf{MESSAGE}$ denotes differentiable message generation function and it can be in various ways (i.e. Multi Layer Perceptrons), and obtain the embedding vector of each pair of the nodes and edge; $\textbf{AGGR}$ denotes the aggregation function (i.e. sum, mean, max) which is differentiable permutation invariant; $\textbf{UPDATE}$ denotes the state update function which usually conducts bias term adding, linear transformation or multi-head processing (i.e. concatenating) to update what we obtain from aggregation process. 

In short, this framework has two steps: Gather the state message from neighbour nodes, and then generate the current node’s message by applying the specific aggregation function; based on the message so far, update the state of the target node.

In our work, the modification can be shown in two parts. In the $\textbf{MESSAGE}$ part, we have
$$\mathbf{\Theta}\mathbf{x}_j \rightarrow (\mathbf{\Theta}\mathbf{x}_j) \, \Vert \, \mathbf{e_{ij}}$$, where $\mathbf{\Theta}$ is the weight matrix and $" \, \Vert \,"$ means concatenation. We concatenate the edge representation $\mathbf{e_{ij}}$ to each corresponding node $\mathbf{x}_j$, and then in the $\mathbf{UPDATE}$ part, an edge update matrix will be multiplied to get the dimension we expect (i.e. the number of relationship types). 

The graph-based models we used in this work: Graph Convolutional Networks (GCN) \cite{kipf2016semi}, Graph Attentional Networks (GAT) \cite{velivckovic2017graph},  Simple Graph Convolutional Networks (SGCN) \cite{wu2019simplifying}, Attention-based Graph Neural Networks (AGNN) \cite{thekumparampil2018attention}, and a special one, Relational Graph Convolutional Networks (RGCN) \cite{schlichtkrull2018modeling}. Noticed that RGCN does not need to be modified since it already considered the relationship between two entities, we put it here as it belongs to one kind of graph-based model. 

\subsection{L-Graph}\label{Sequence-based} Instead of conditioning the supporting facts to represent the query in the graph-based model, we process the facts and query independently. Inputs in the sequence-based encoder are the subject-predicate-object (SPO) triples by linearising a relational graph that extracted from supporting facts within a story, as in \cite{minervini2020learning}. Actually supporting facts within a story can be seen as a reasoning path, we only pay attention to nodes(i.e. entities) and edges (i.e. relationship) among them. Therefore, subject and object correspond to two entities, and a predicate is a relationship between these two entities. After extracting all SPO triples in a story, we then directly put them (i.e. SPO sequence) into standard sequence models (i.e. CNN, RNN), since they are ordered and can be seen as a short/key sentence compared to the original story. Similarly, we represent the query into the SO sequence which similar to SPO sequence but without the predicate/target (i.e. relationship) and then put it into sequence models as well. At last, after concatenating and softmax function, we can predict the relationship by choosing the highest value.

To be specific. The sequence-based model does not process query conditioning on the supporting facts, in contrast, supporting facts and query are handled independently:
$$\hat{y} = softmax\left(\mathbf{W}[emb(\mathcal{F}) \, \Vert \, emb(\mathcal{Q})]\right)$$
where $"emb"$ denotes an embedding process working on a set of ground facts, such that the supporting facts ($\mathcal{F}$) or a query ($\mathcal{Q}$); $"\, \Vert \,"$ means a concatenation of the two items; $\mathbf{W}$ represents weight matrix; and $\hat{y}$ represents the distribution of all possible relationships we get after softmax function, and we will choose the highest probability relationship type as the predicted target.

Here, $emb(\mathcal{F})$ is from the result of processing SPO sequences with sequence encoding models, such as CNN and RNN, while $emb(\mathcal{Q})$ without predicate in the processing inputs (i.e. SO sequences). The shape of them both are [$B \, \times \, emb_{dim}^*$], where $B$ is the batch size, $emb_{dim}^*$ is the embedding dimension for each story (consist of SPO sequences and SO sequences) based on sequence models, and [$B \, \times \, (2*emb_{dim}^*)$] after concatenation. For example, when the LSTM \cite{Hochreiter_Schmidhuber_1997} is used with hidden size = 100, we can get [$B \, \times \, 100$] for both processes and get [$B \, \times \, 200$] after concatenating, and if we use the Bidirectional Long Short-Term Memory Networks (Bi-LSTM) \cite{thireou2007bidirectional}, $emb_{dim}^*$ will become twice of that in the LSTM.

We consider several sequence encoding models, namely Recurrent Neural Networks (RNN) \cite{Rumelhart_Hinton_Williams_1986}, Long Short-Term Memory Networks (LSTM) \cite{Hochreiter_Schmidhuber_1997},
Gated Recurrent Units (GRU) \cite{cho2014learning}, Bidirectional Recurrent Neural Networks (Bi-RNN) \cite{schuster1997bidirectional}, Bidirectional Long Short-Term Memory Networks (Bi-LSTM) \cite{thireou2007bidirectional}, Bidirectional Gated Recurrent Units (Bi-GRU) \cite{vukotic2016step}, Convolutional Neural Networks (CNN) \cite{LeCun1989}, CNN with Highway Encoders (CNNH) \cite{kim2015character}, Multi-Headed Attention Networks (MHA) \cite{vaswani2017attention}. All related codes can be easily built up based on "torch.nn" or the NLP research library, named "AllenNLP" \cite{Gardner2017AllenNLP}. 

\section{Experimental Results \label{char4}}
We build up several baselines for evaluating the systematic generalisation and robust reasoning capacity on the pre-generated CLUTTR datasets. Models' performance is comparing to each other, including "text-based" models \cite{sinha2019clutrr} mentioned in the paper. The prefix "graph\_" using in the sequence-based model (L-Graph) is to distinguish from the text-based model.

\subsection{Systematic Generalisation Evaluation}
We have conducted experiments on two pre-generated CLUTRR datasets, "data\_ 089907f8" and "data\_db9b8f0", to compare each models' systematic generalisation capacity. Due to limited pages, only partly results under maximising validation accuracy metric based on "data\_db9b8f0" are illustrated. Numerical details for all models under both minimising validation loss metric and maximising validation accuracy metric did not show here, as well as optimal hyperparameters.

\begin{figure}[H]
\centering
\includegraphics[width=1\textwidth]{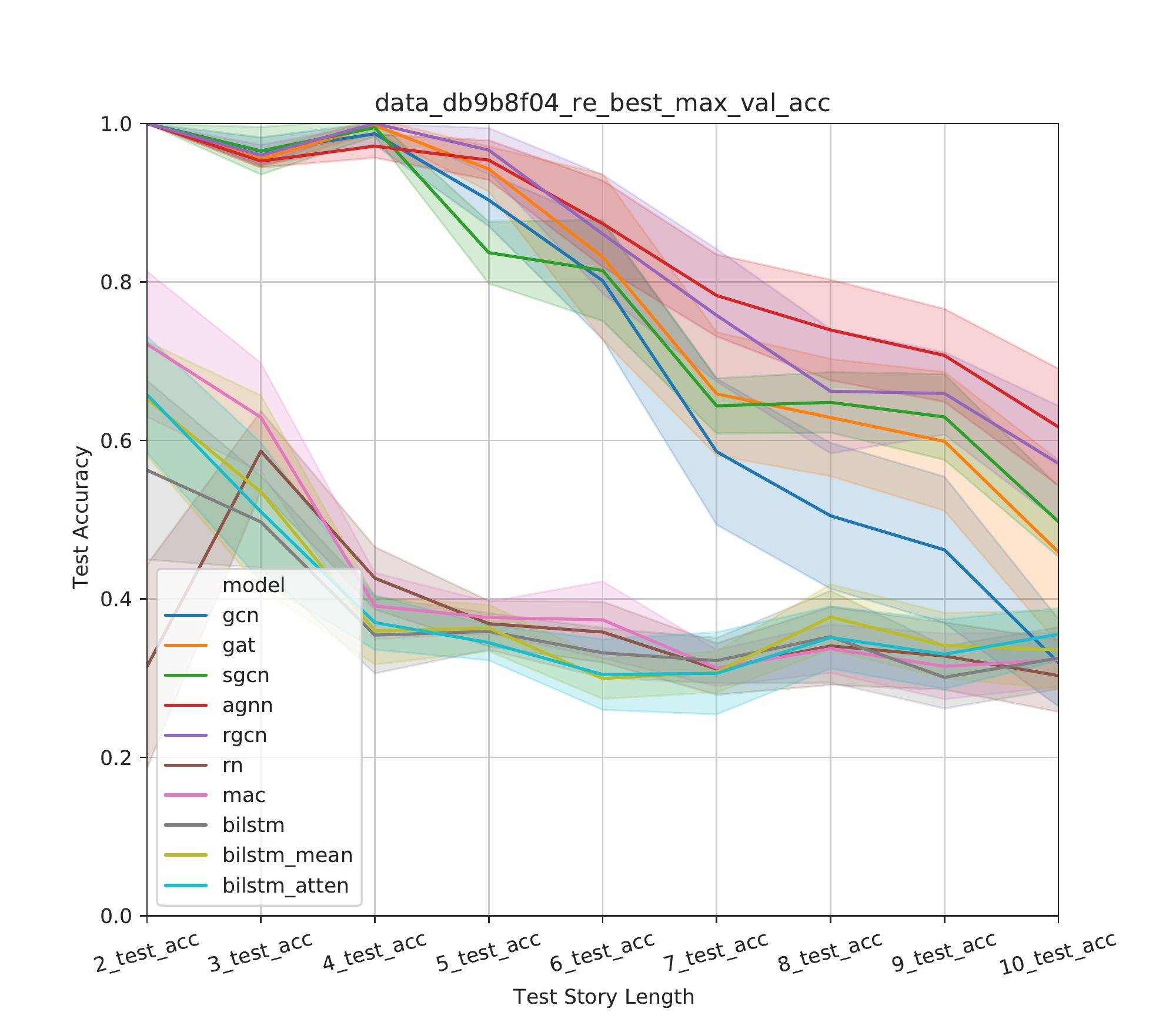}
\caption{Systematic generalisation performance of graph-based models when trained on clauses of length k = 2, 3, 4.\\
In ascending order, the mean test accuracy of graph-based models are: 0.7255 (gcn), 0.7812 (sgcn), 0.7858 (gat), 0.8258 (rgcn) and 0.8442 (agnn); the test accuracy of clauses length $k$ = 10 are: 0.3196 (gcn), 0.4591 (gat), 0.4975 (sgcn), 0.571 (rgcn) and 0.617 (rgcn).}
\label{Fig.1.2} 
\end{figure}

The gap between two types of model is quite evident in \textbf{Fig. \ref{Fig.1.2}}. Graph-based models have more than 0.9 accuracies on clauses length 2, 3 and 4, and then drop significantly till clauses length 10. Among them, AGNN can be regarded as the best model; it obtains the highest value across most of the situations and even at length 10 with a figure of 0.617. Different from the monotonic decline in the graph-based models, over the length from 4 to 10, the text-based models' testing accuracy remained level. Although the graph-based model performs much better than text-based in most scenarios, they decrease very fast and at most 0.6 accuracies can be obtained at length 10 and the worst one only about 0.3.

\begin{figure}[H]
\centering
\includegraphics[width=1\textwidth]{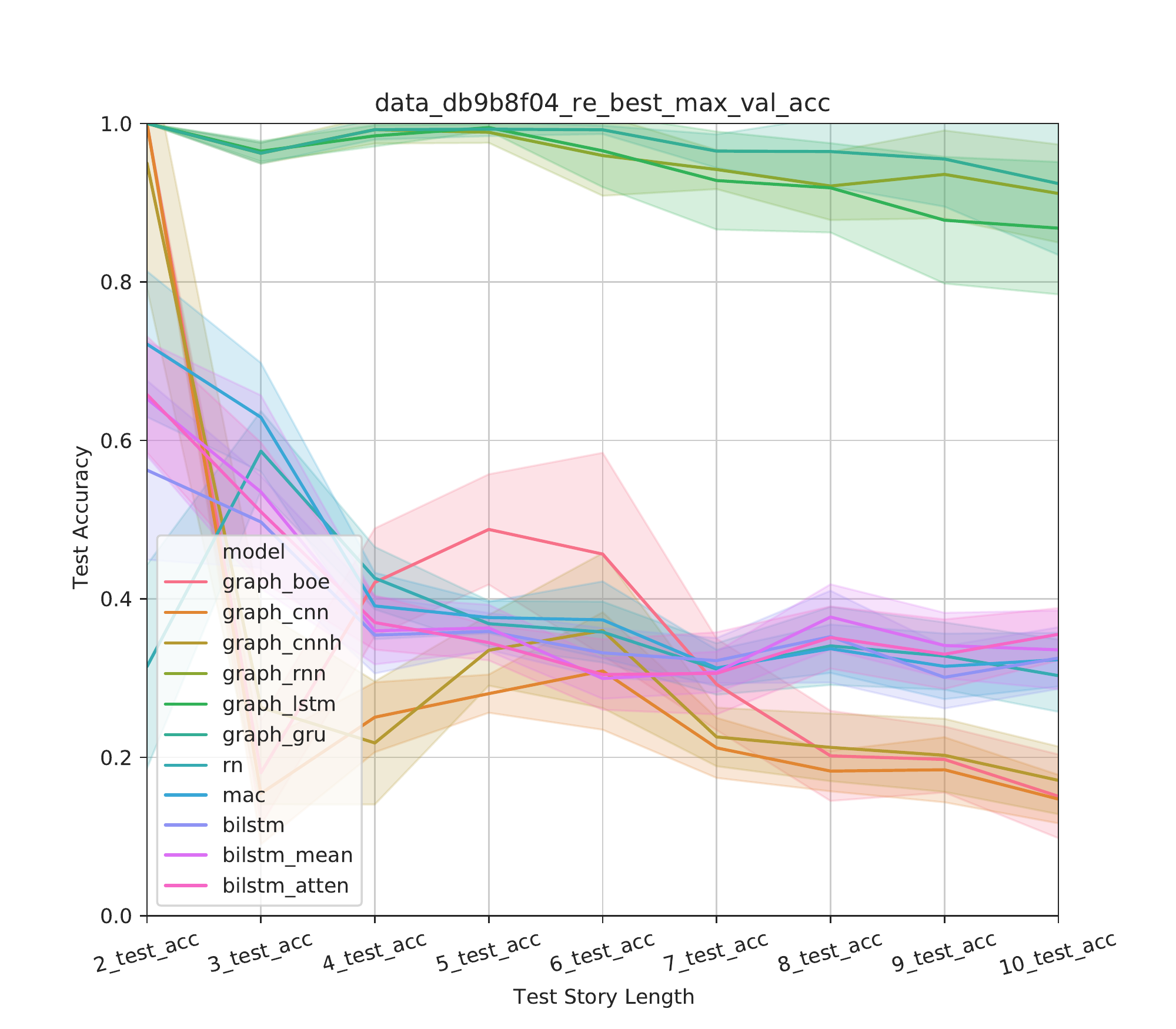}
\caption{Systematic generalisation performance of common sequence models when trained on clauses of length k = 2, 3, 4.\\ 
In ascending order, the mean test accuracy of sequence-based models are: 0.3023 (graph\_cnn), 0.3265 (graph\_cnnh), 0.3764 (graph\_boe), 0.9447 (graph\_lstm), 0.9571 (graph\_rnn) and 0.9721 (graph\_gru); the test accuracy of clauses length $k$ = 10 are: 0.1473 (graph\_cnn), 0.1509 (graph\_boe), 0.171 (graph\_cnnh), 0.8679 (graph\_lstm), 0.9116 (graph\_rnn) and 0.9241 (graph\_gru).}
\label{Fig.2.2} 
\end{figure}

RNN, LSTM and GRU perform much stable with high test accuracy in \textbf{Fig. \ref{Fig.2.2}} (trained on clauses of length $k$ = 2, 3, 4), with nearly 0.9 test accuracy on clauses length 10 stories. However, BOE, CNN and CNNH decline in an unbelievable speed, from nearly 1.0 test accuracy on clauses length 2 to only about 0.1 on clauses length 3, while test-based models do not show much difference. 

\subsection{Robust Reasoning Evaluation} 
We also have conducted experiments on four pre-generated CLUTRR datasets, "data\_7c5b0e70", "data\_06b8f2a1", "data\_523348e6" and "data\_d83ecc3e", to compare each models' robust reasoning capacity. Numerical details for all models under both minimising validation loss metric and maximising validation accuracy metric did not show here, and optimal hyperparameters as well.

\section{Conclusions}
In this paper, we implemented and evaluated two neuro-symbolic reasoning architectures for testing systematic generalisation and inductive reasoning capabilities of NLU systems, namely the E-GNN (Graph-based) model and the L-Graph (Sequence-based) model. Both two types of models were trained on the CLUTRR datasets, and we demonstrated and evaluated quite a lot of experiments results in both generalisation and robustness aspects. 

Most models can perform well with fewer clauses length inputs, however, as the length of clauses increasing in the test stage, all models' performance decline monotonically, especially for graph-based models, which proves the challenge of "zero-shot" systematic generalisation \cite{lake2018generalization,sodhani2018training}. Among them, those sequence-based models with recurrent neural networks outperform the graph-based model, the typical sequence model, and the multi-head attention networks model when considering generalisation capacity, with around 90\% on clauses length 10 when trained over clauses of length 2, 3 and 4. Despite that, their robustness performance cannot as good as that of graph-based models and sometimes even fail to obtain any correct kinship. 

These results highlight that graph-based models with attention architectures can catch the patterns of the data and prevent from diverse noise influence; sequence-based models can make good use of SPO sequences to predict the kinship between entities; however, we can also notice that there is a fatal flaw in this strategy---very sensitive to the order of sequences; bidirectional settings and multi-heads attention architecture did not have much more advantage than the original ones; although models were trained on data with noise added, most of them may be able to get some linguistic cues (i.e. gender) and predict well under that type of noise scenarios on clauses of length $k$ = 2, but testing on clauses of length $k$ = 3 is still a challenge for some models.

Structured input plays a significant role in all those models. Compared with text-based models, the performance of models with structured inputs improves significantly, especially for models with recurrent neural networks, such as LSTM and GRU, show us an unbelievable accuracy during the systematic generalisation tests. Moreover, this phenomenon also proves that there is a gap between reasoning models trained on structured input or unstructured natural language.

\bibliographystyle{splncs04}
\bibliography{paper}
%





\newpage
\appendix
\section{Kinship and Knowledge Base}
\label{kb}

Table \ref{Table.kinship} is the kinship used in the CLUTRR. Table \ref{Table.kb} is the Knowldge Base ($\mathcal{KB}$) used in the CLUTRR for generating the datasets, where the prefix "inv-" means the "parent-to-child" relation, while the others are the "child-to-parent" relation. For example, we can use "$[inv-grand, X, Y]$" to represent "$X$ is the grandfather of $Y$", while "$[grand, X, Y]$" means "$X$'s grandfather is $Y$" or "$Y$ is the grandfather of $X$" with the same meaning. 

\begin{table}[H]
\caption{Kinship used in the CLUTRR}
\label{Table.kinship}
\centering
\includegraphics[width=1\textwidth]{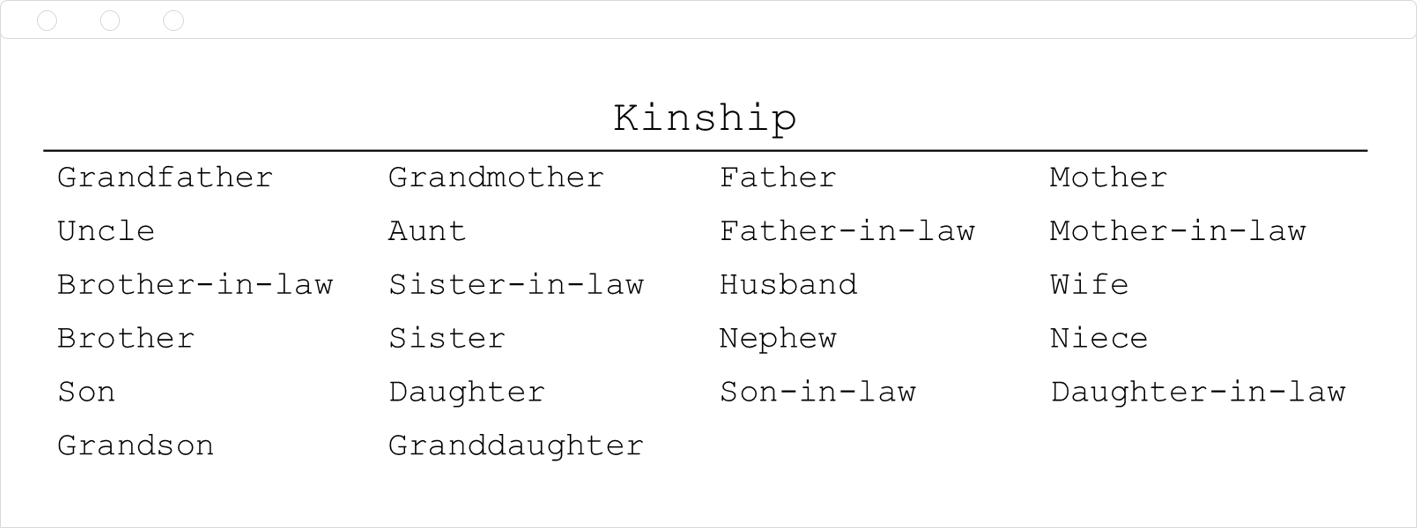}
\end{table}

\begin{table}[H]
\caption{Knowledge Base ($\mathcal{KB}$) used in the CLUTRR}
\label{Table.kb}
\centering
\includegraphics[width=1\textwidth]{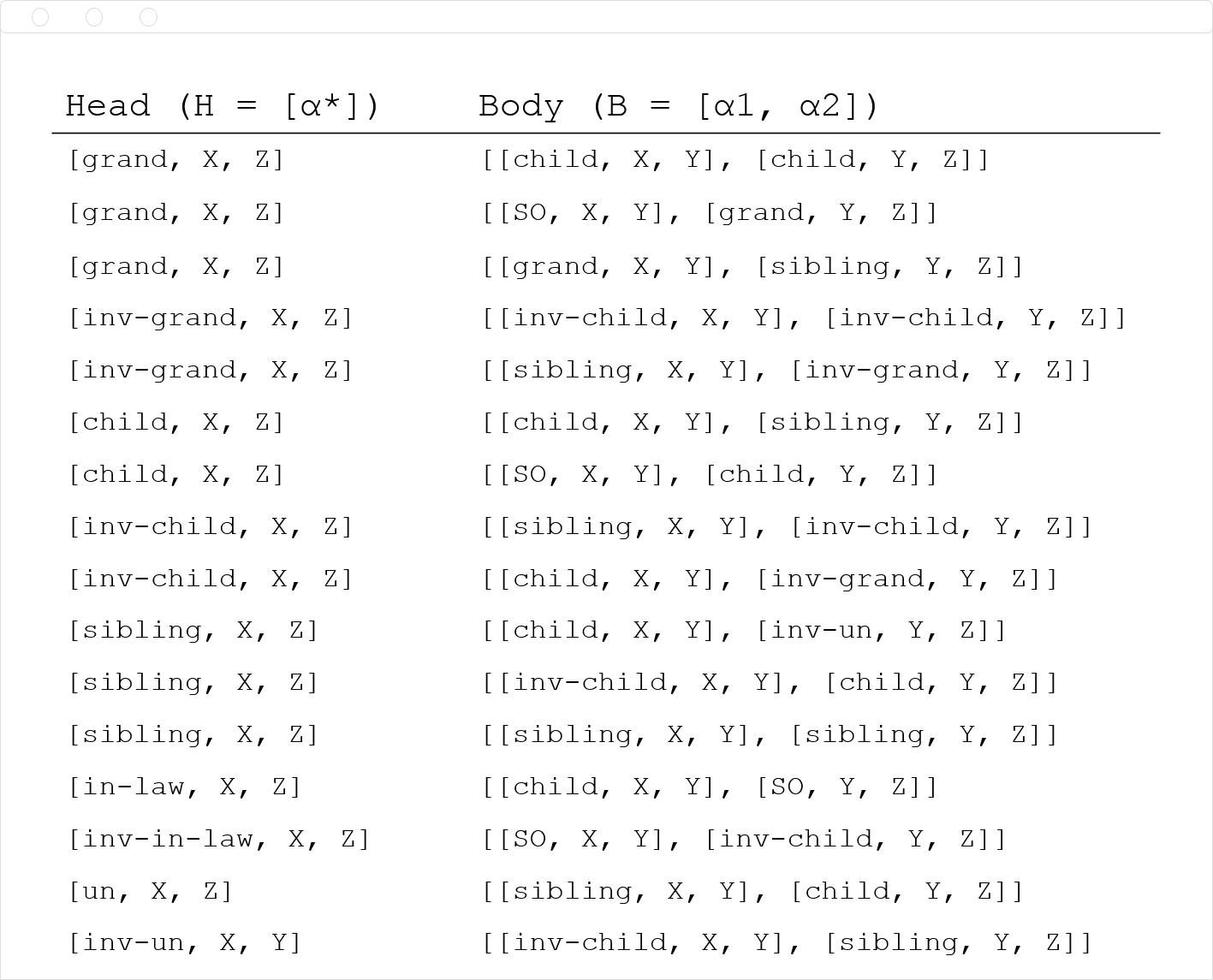}
\end{table}

\section{Datasets}
\label{Datasets}

\begin{table}[H]
\caption{Details of pre-generated datasets in the CLUTRR}
\centering
\includegraphics[width=0.8\textwidth]{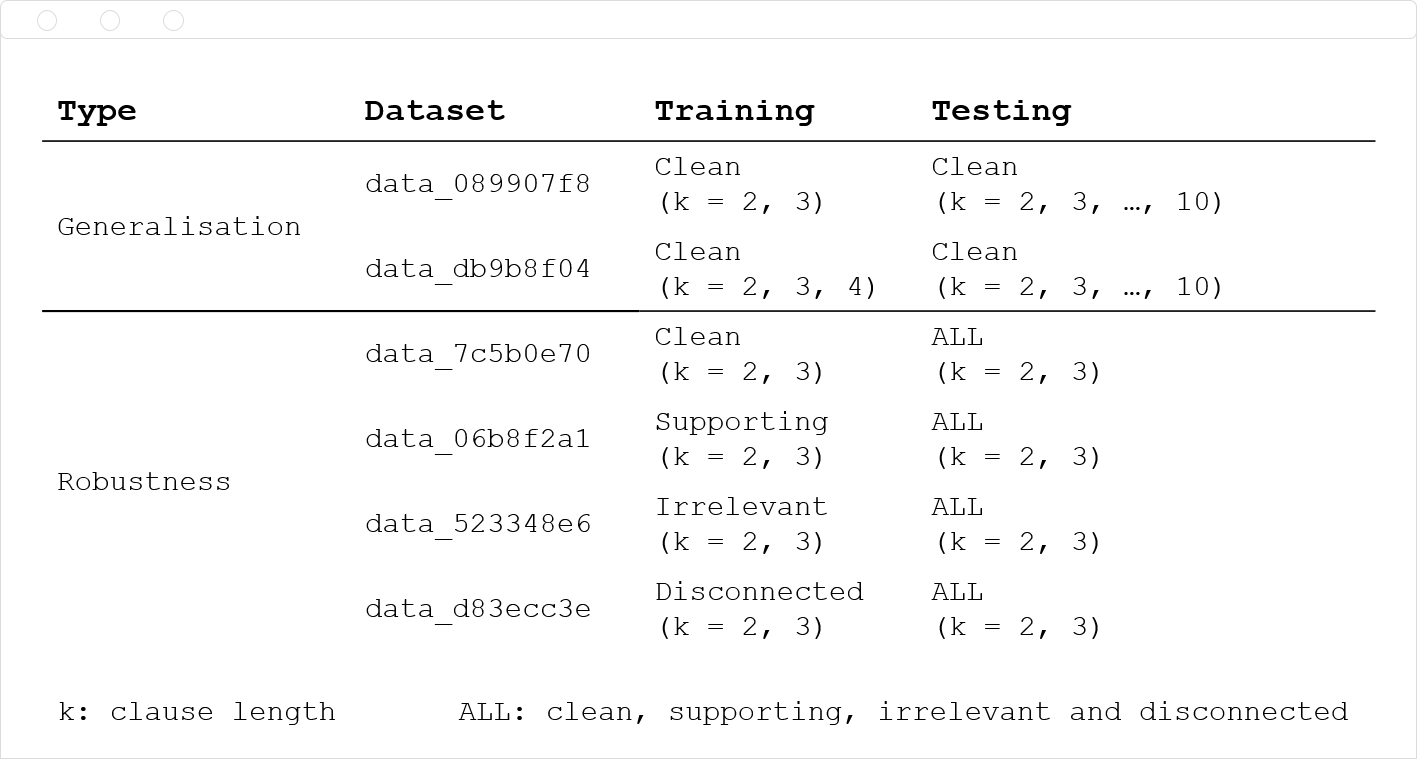}
\end{table}

\begin{table}[H]
\caption{An example in the CLUTRR datasets}
\centering
\includegraphics[width=0.9\textwidth]{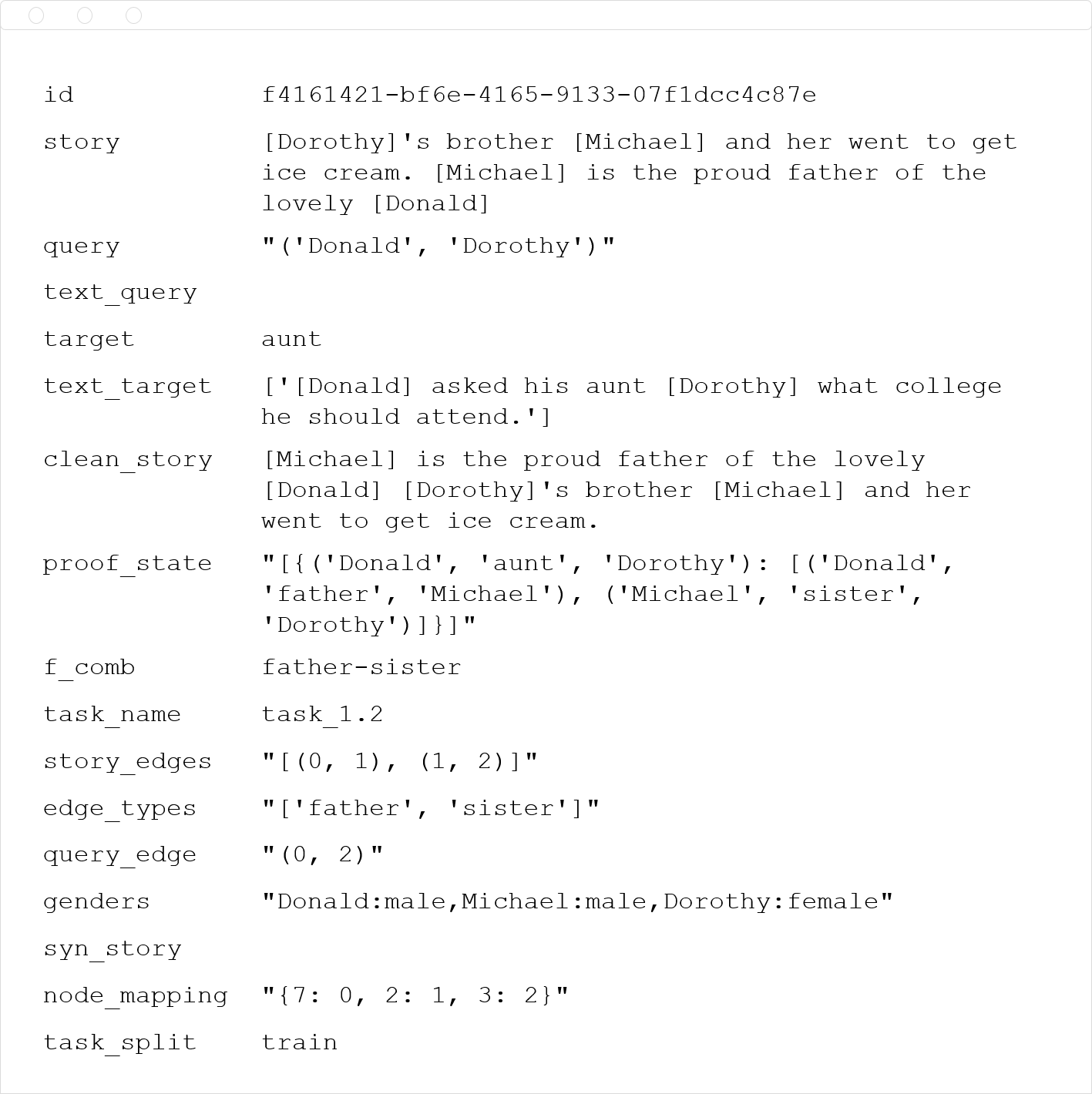}
\end{table}

\section{Abbreviations}
\label{abb}
For brevity, we use some abbreviations to represent hyperparameters and models in our work and code. The prefix "graph" using in the sequence-based model is to distinguish from the text-based model.

\begin{table}[H]
\caption{The Abbreviations in this work}
\centering
\includegraphics[width=1\textwidth]{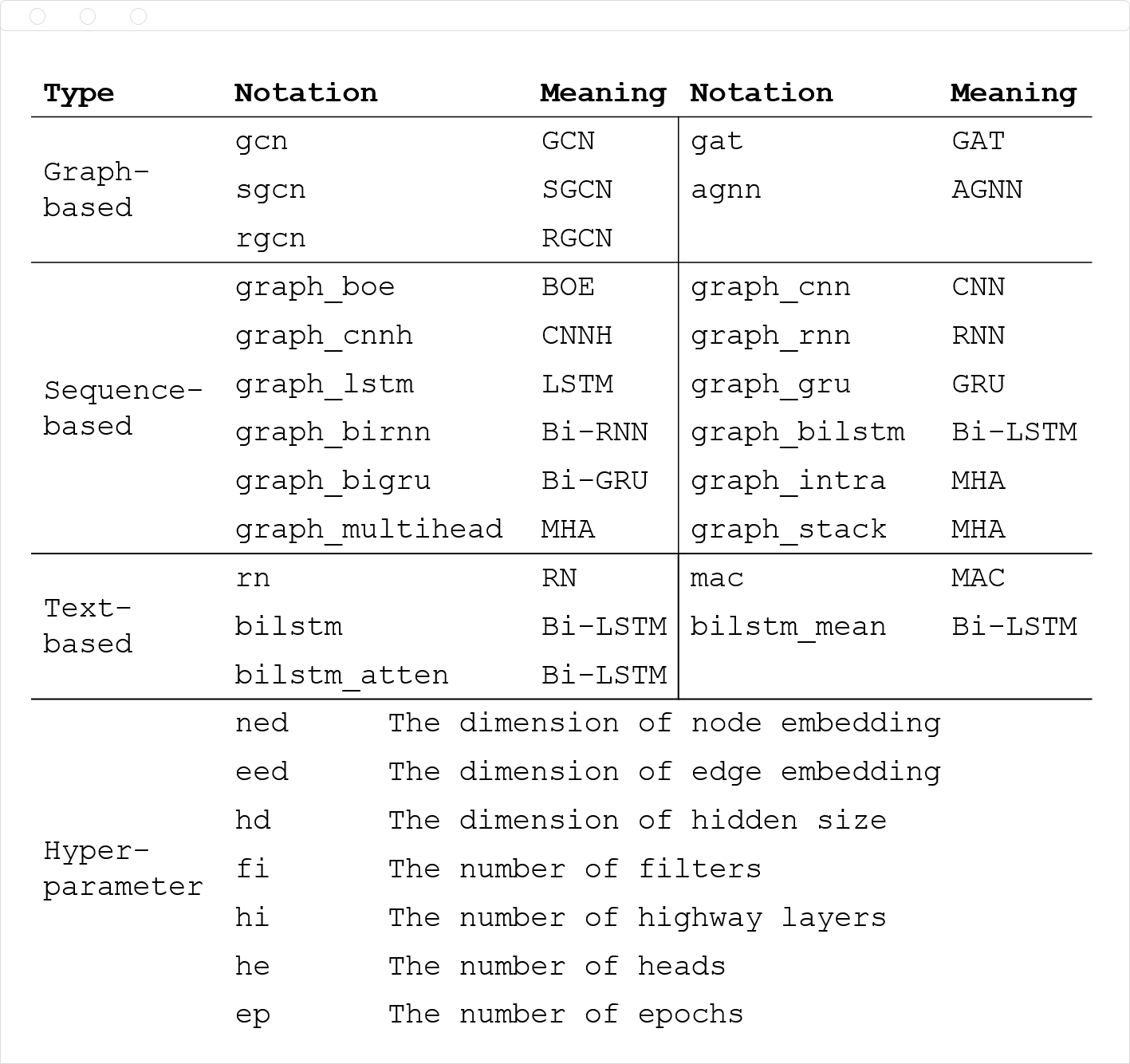}
\end{table}

\section{Systematic Generalisation}
\label{sys}

\subsection{Clauses of Length k = 2, 3}
\label{k23}

\begin{table}[H]
\caption{Optimal hyperparameters for training on clauses of length k = 2, 3 under minimum validation loss metric (left) and maximum validation accuracy metric (right).}
\label{Table.gh1}
\center
\includegraphics[width=\textwidth]{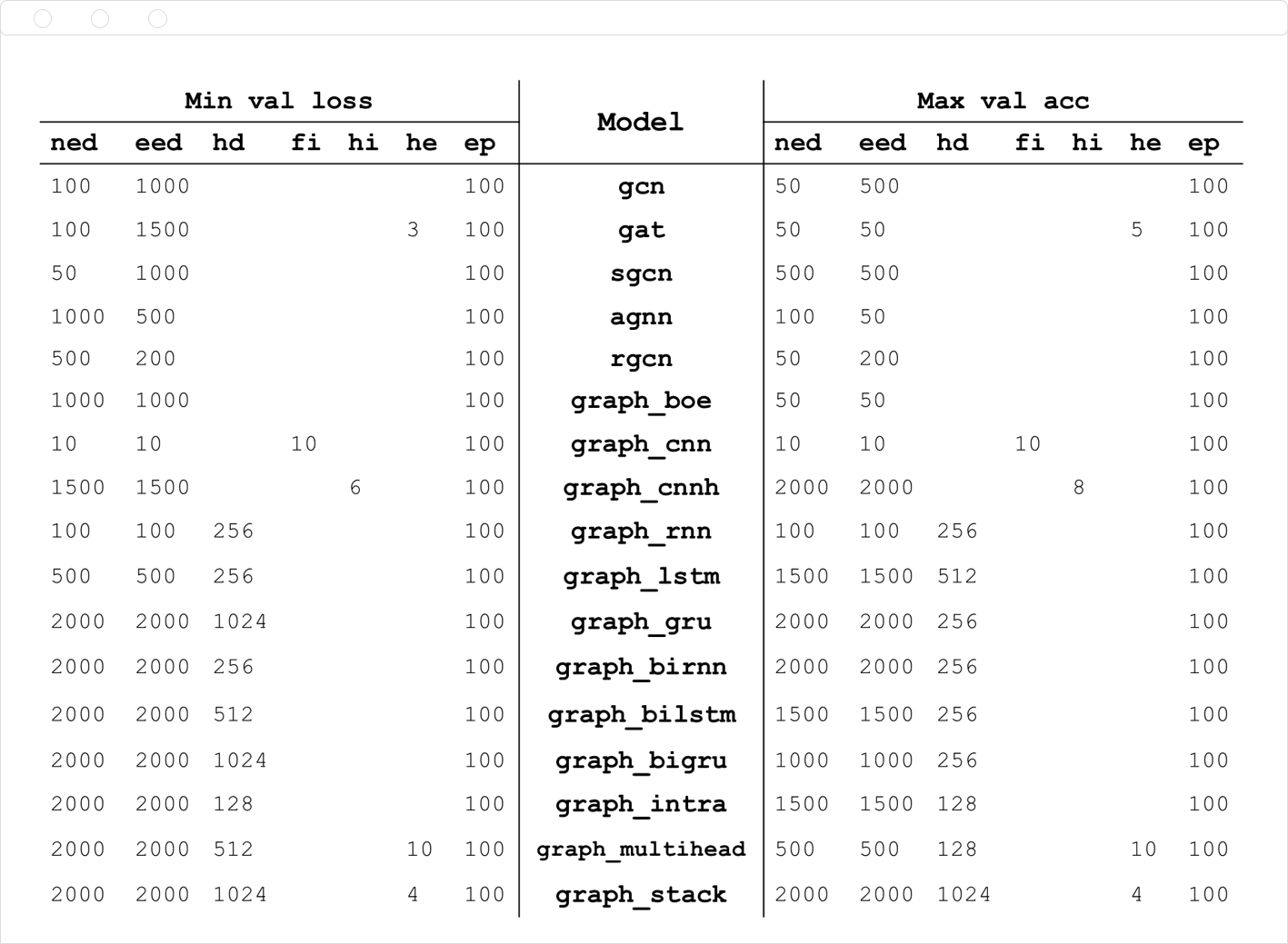}
\end{table}

\newpage
\begin{table}[H]
\caption{Training on clauses of length k = 2, 3 under minimum validation loss metric. The optimal values within the graph-based, the sequence-based and the text-based are in bold; the best performance among all models are in red.}
\label{Table.g11}
\center
\includegraphics[width=1\textwidth]{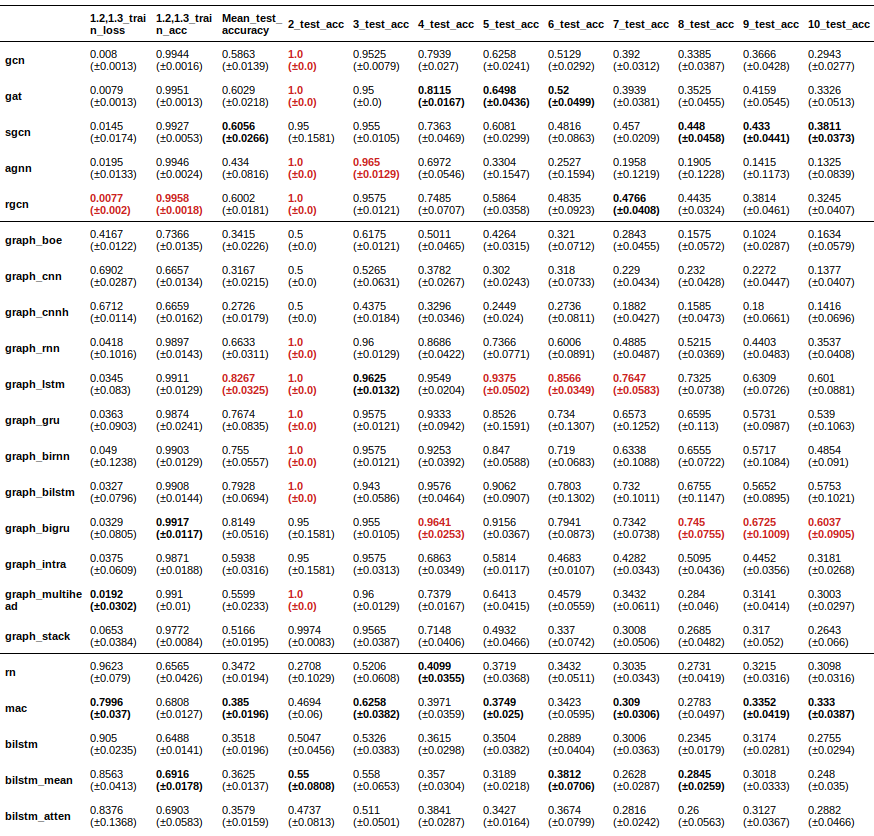} 
\end{table}

\newpage
\begin{table}[H]
\caption{Training on clauses of length k = 2, 3 under maximum validation accuracy metric. The optimal values within the graph-based, the sequence-based and the text-based are in bold; the best performance among all models are in red.}
\label{Table.g12}
\center
\includegraphics[width=1\textwidth]{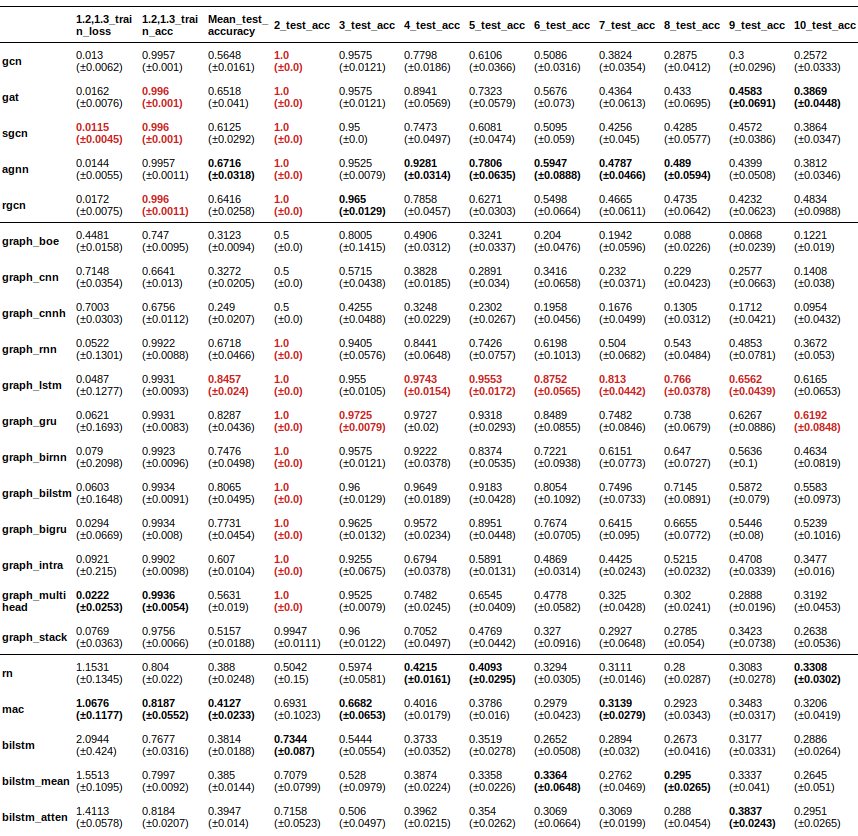} 
\end{table}

\subsection{Clauses of Length k = 2, 3, 4}
\label{k234}

\begin{table}[H]
\caption{Optimal hyperparameters for training on clauses of length k = 2, 3, 4 under minimum validation loss metric (left) and maximum validation accuracy metric (right).}
\label{Table.gh2}
\center
\includegraphics[width=\textwidth]{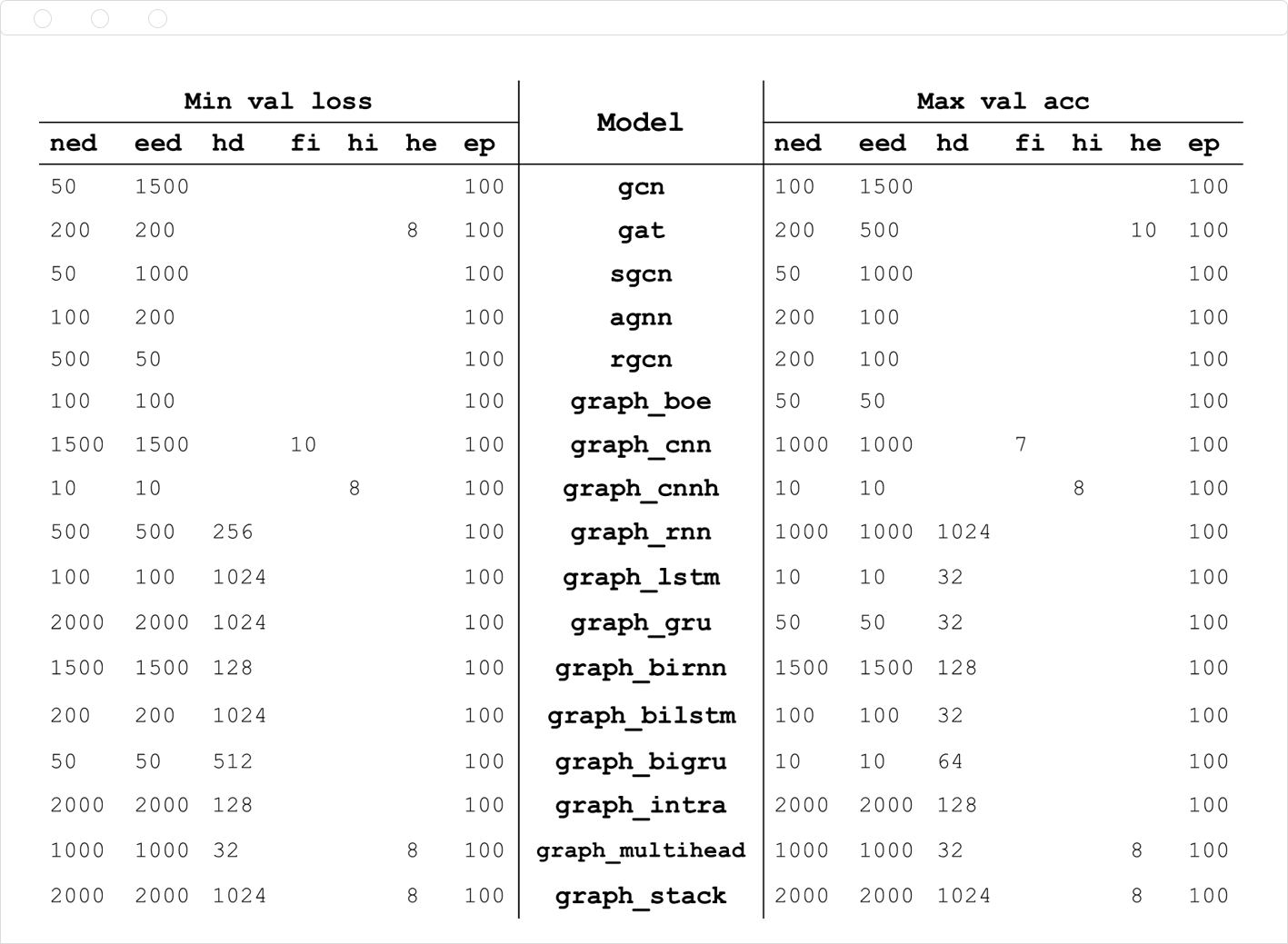}
\end{table}

\newpage
\begin{table}[H]
\caption{Training on clauses of length k = 2, 3, 4 under minimum validation loss metric. The optimal values within the graph-based, the sequence-based and the text-based are in bold; the best performance among all models are in red.}
\label{Table.g21}
\center
\includegraphics[width=1\textwidth]{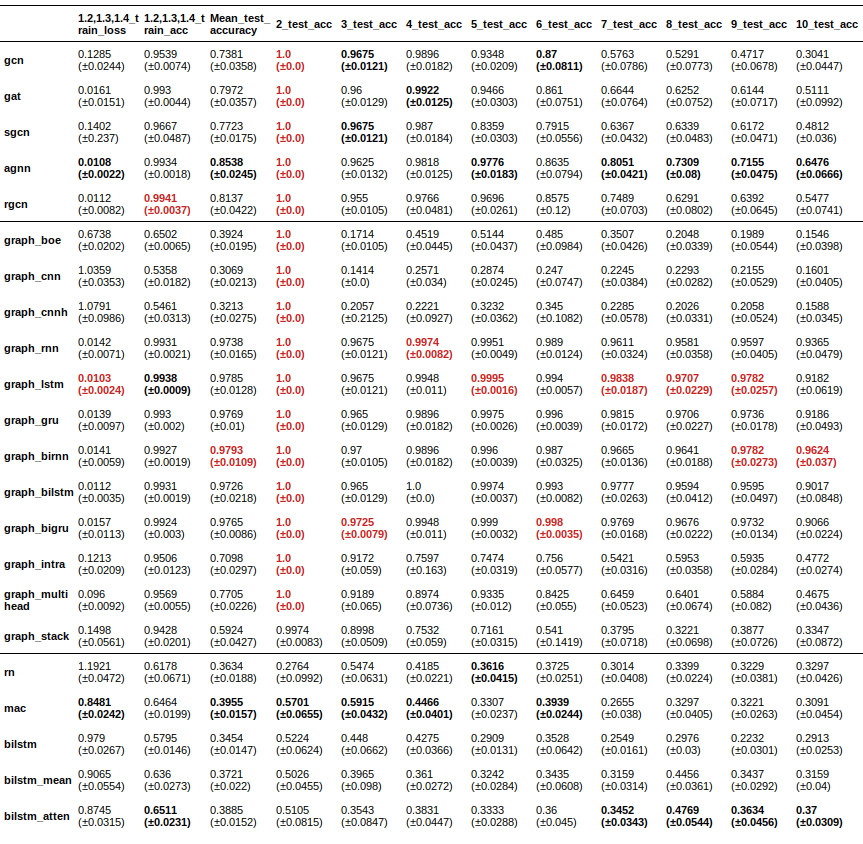} 
\end{table}

\newpage
\begin{table}[H]
\caption{Training on clauses of length k = 2, 3, 4 under maximum validation accuracy metric. The optimal values within the graph-based, the sequence-based and the text-based are in bold; the best performance among all models are in red.}
\label{Table.g22}
\center
\includegraphics[width=1\textwidth]{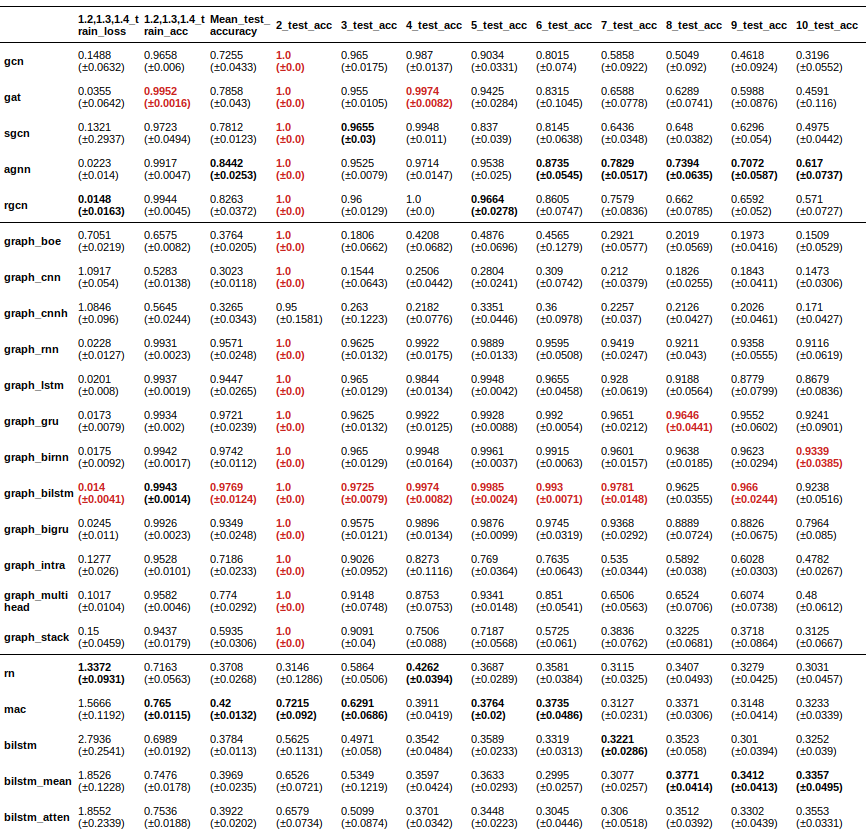} 
\end{table}

\section{Robust Reasoning}
\label{rob}

\subsection{Clean Facts without Noise}
\label{Clean}

\begin{table}[H]
\caption{Optimal hyperparameters for training on the clean facts without noise under minimum validation loss metric (left) and maximum validation accuracy metric (right).}
\label{Table.gh3}
\center
\includegraphics[width=\textwidth]{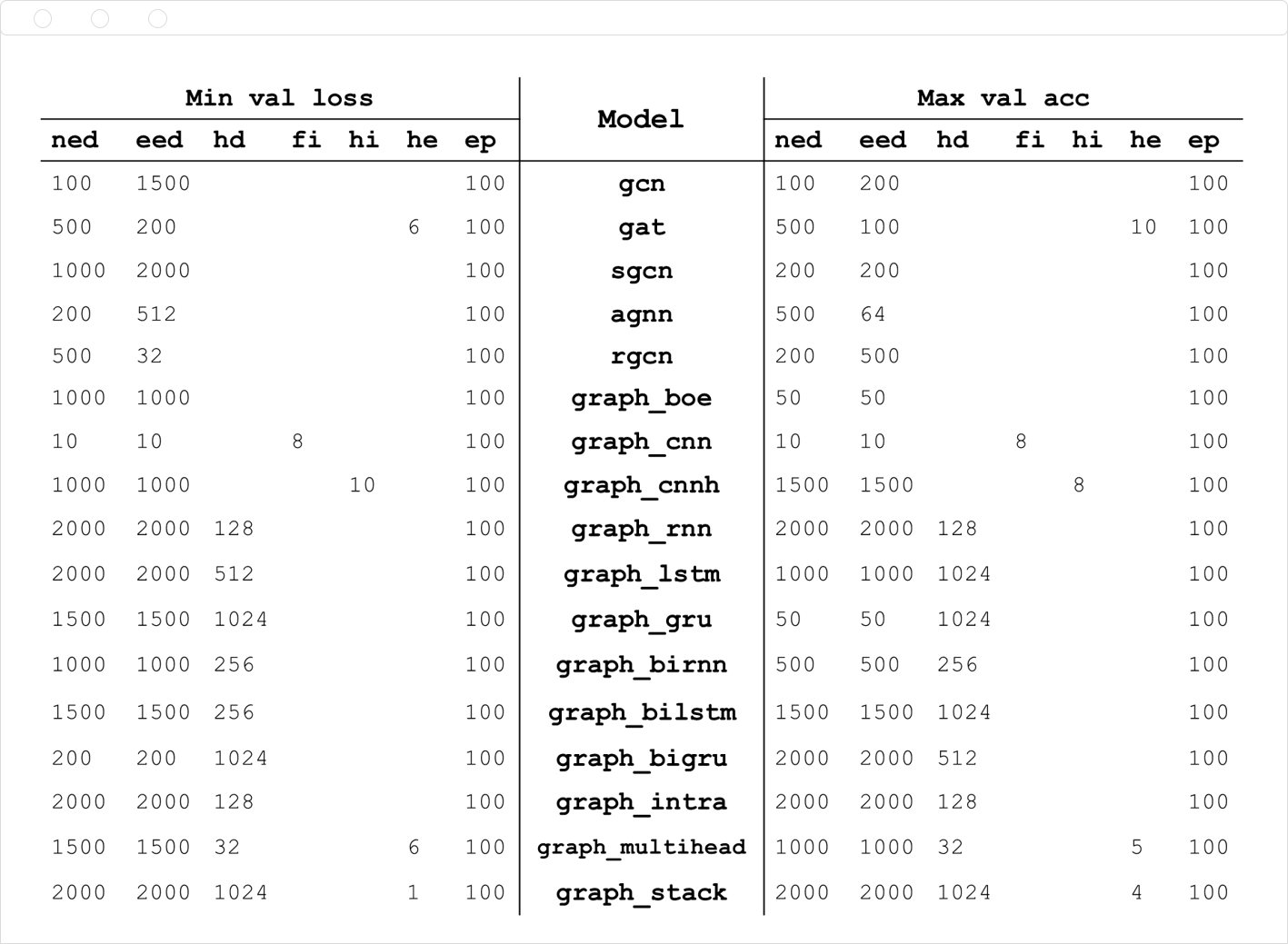}
\end{table}

\newpage
\begin{table}[H]
\caption{Training on the clean facts without noise under minimum validation loss metric. The optimal values within the graph-based, the sequence-based and the text-based are in bold; the best performance among all models are in red.}
\label{Table.g31}
\center
\includegraphics[width=0.98\textwidth]{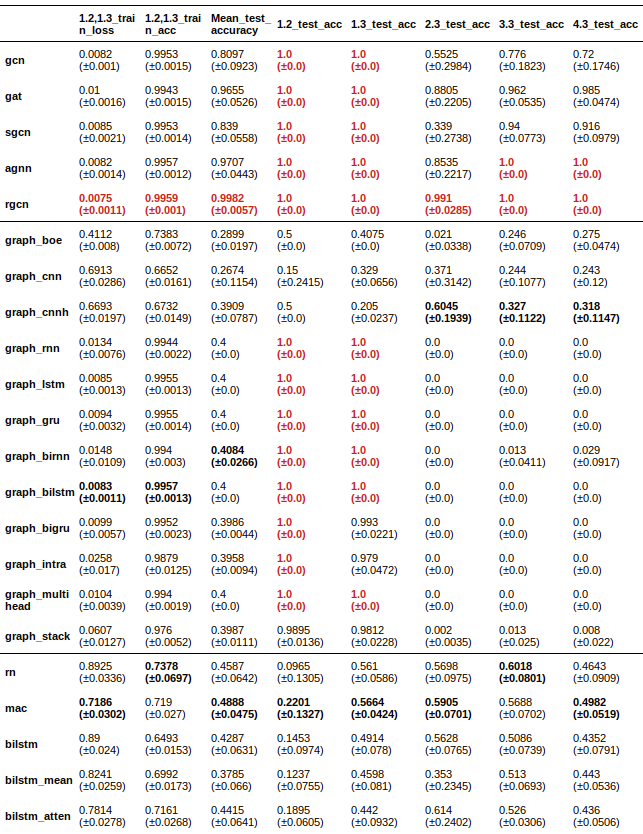} 
\end{table}

\newpage
\begin{table}[H]
\caption{Training on the clean facts without noise under maximum validation accuracy metric. The optimal values within the graph-based, the sequence-based and the text-based are in bold; the best performance among all models are in red.}
\label{Table.g32}
\center
\includegraphics[width=0.98\textwidth]{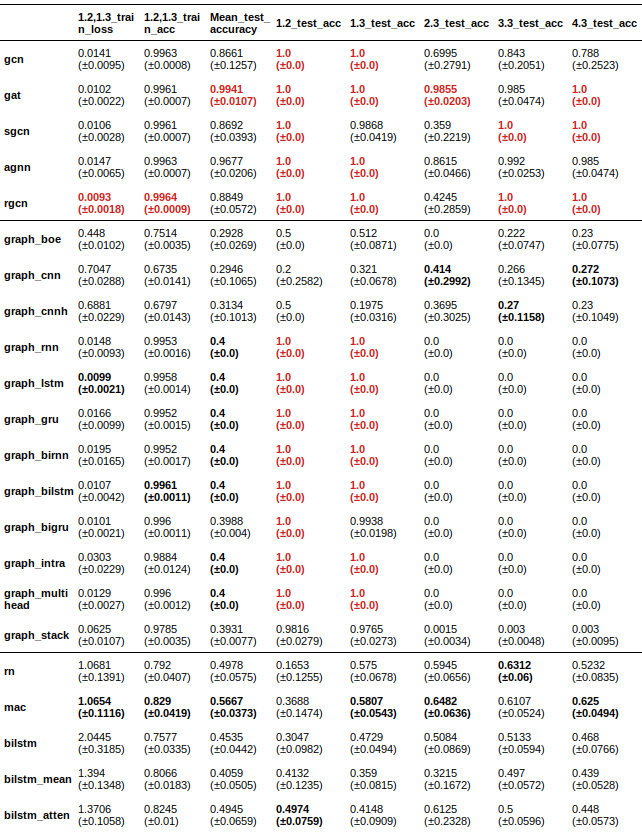} 
\end{table}

\subsection{Supporting Facts Noise}
\label{Supporting}

\begin{table}[H]
\caption{Optimal hyperparameters for training on the supporting facts noise under minimum validation loss metric (left) and maximum validation accuracy metric (right).}
\label{Table.gh4}
\center
\includegraphics[width=\textwidth]{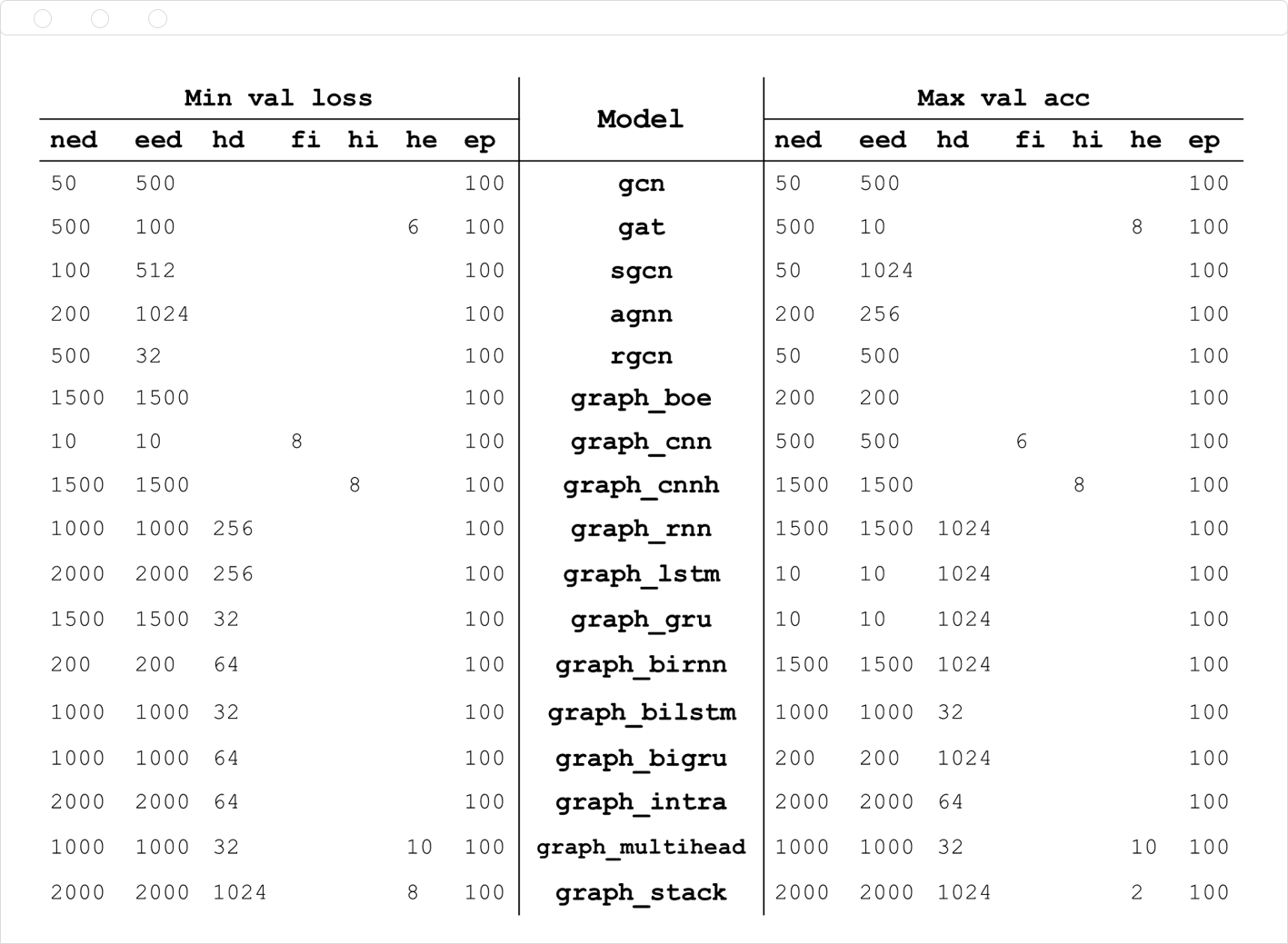}
\end{table}

\newpage
\begin{table}[H]
\caption{Training on the supporting facts noise under minimum validation loss metric. The optimal values within the graph-based, the sequence-based and the text-based are in bold; the best performance among all models are in red.}
\label{Table.g41}
\center
\includegraphics[width=0.98\textwidth]{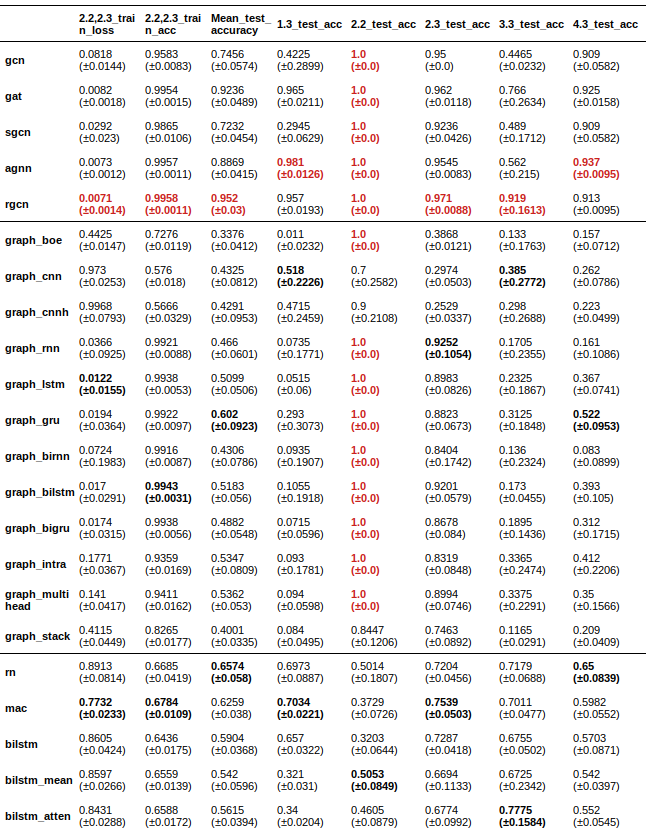} 
\end{table}

\newpage
\begin{table}[H]
\caption{Training on the supporting facts noise under maximum validation accuracy metric. The optimal values within the graph-based, the sequence-based and the text-based are in bold; the best performance among all models are in red.}
\label{Table.g42}
\center
\includegraphics[width=0.98\textwidth]{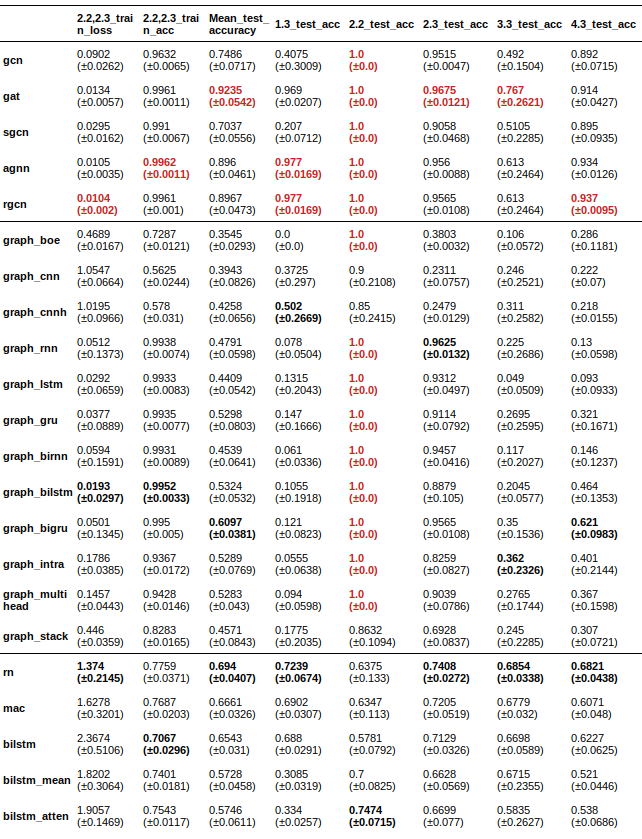} 
\end{table}

\subsection{Irrelevant Facts Noise}
\label{Irrelevant}

\begin{table}[H]
\caption{Optimal hyperparameters for training on the irrelevant facts noise under minimum validation loss metric (left) and maximum validation accuracy metric (right).}
\label{Table.gh5}
\center
\includegraphics[width=\textwidth]{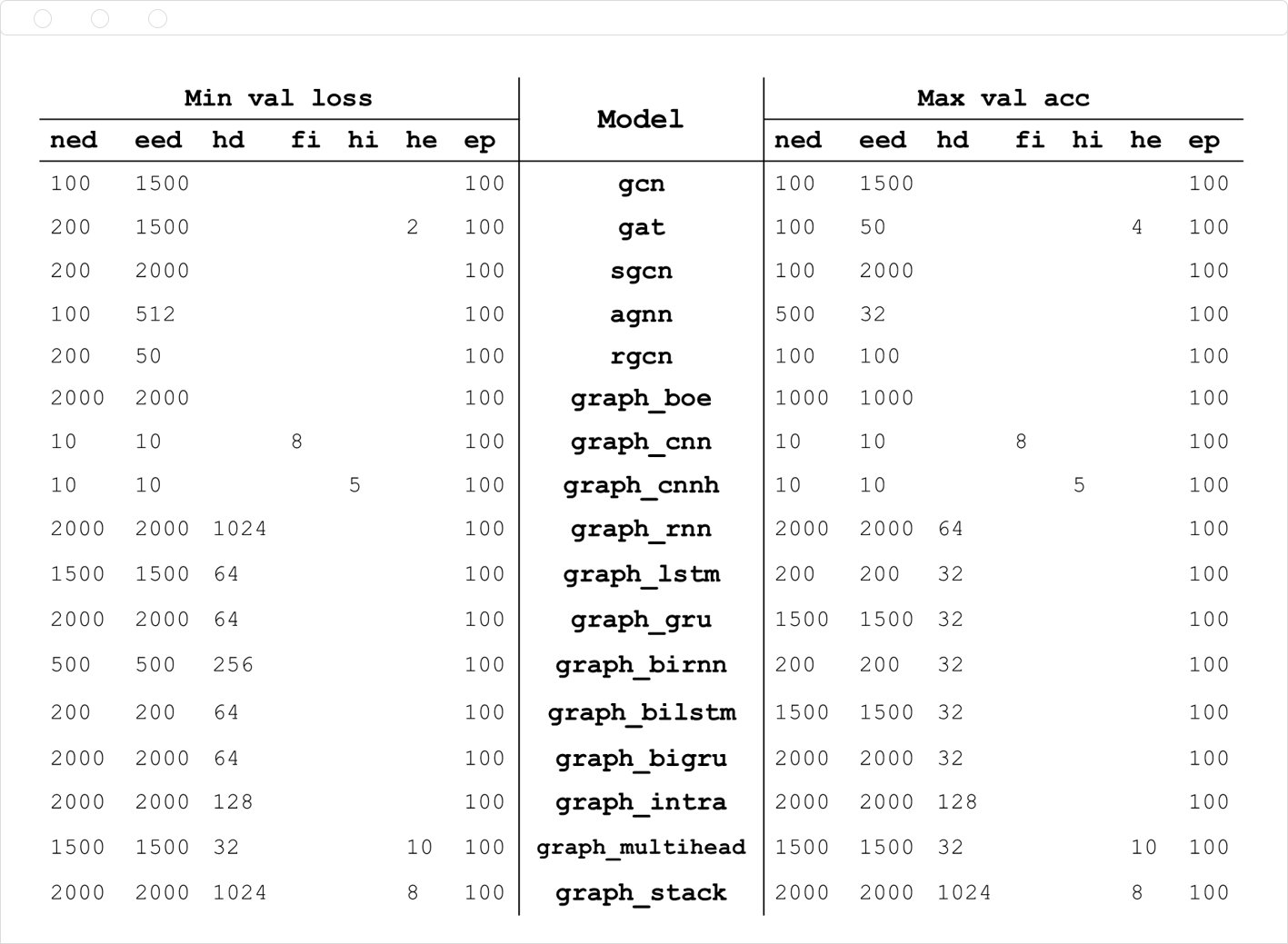}
\end{table}

\newpage
\begin{table}[H]
\caption{Training on the irrelevant facts noise under minimum validation loss metric. The optimal values within the graph-based, the sequence-based and the text-based are in bold; the best performance among all models are in red.}
\label{Table.g51}
\center
\includegraphics[width=0.98\textwidth]{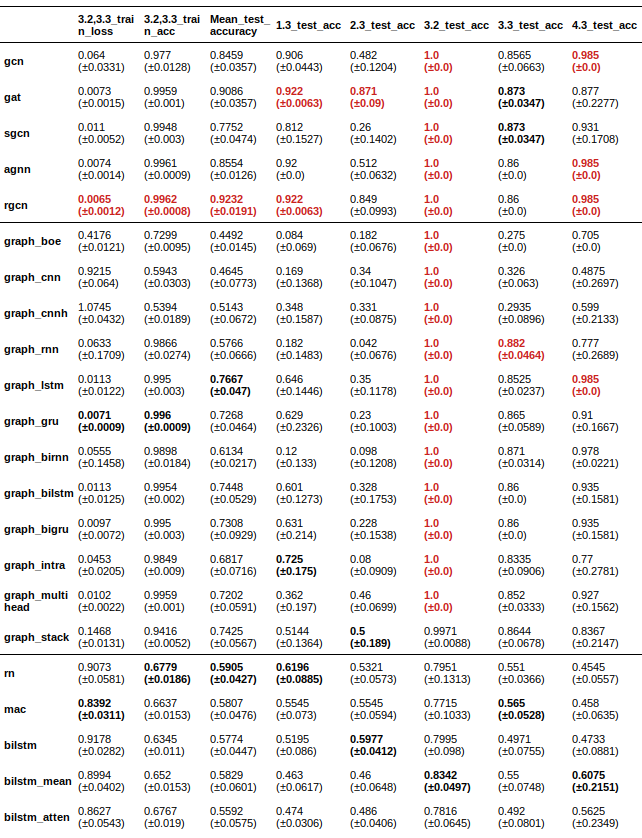} 
\end{table}

\newpage
\begin{table}[H]
\caption{Training on the irrelevant facts noise under maximum validation accuracy metric. The optimal values within the graph-based, the sequence-based and the text-based are in bold; the best performance among all models are in red.}
\label{Table.g52}
\center
\includegraphics[width=0.98\textwidth]{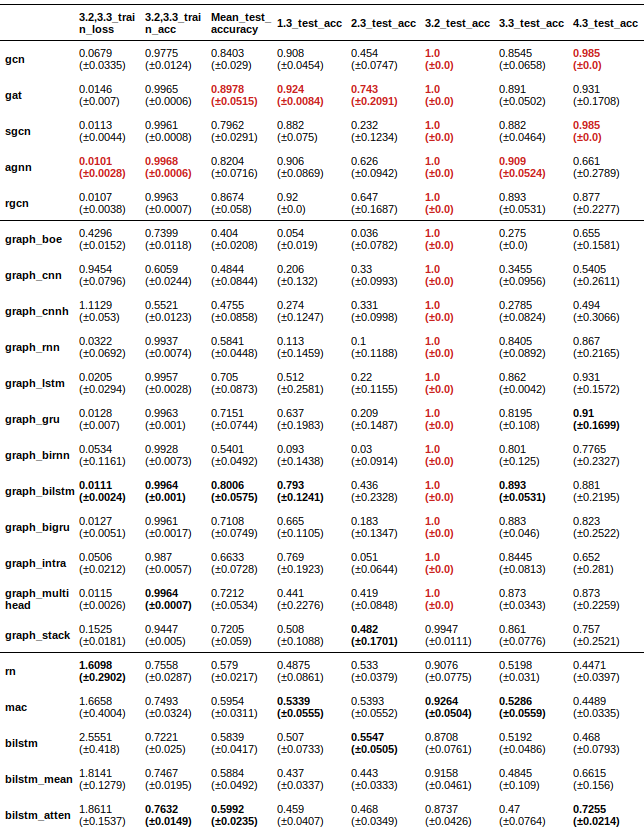} 
\end{table}

\subsection{Disconnected Facts Noise}
\label{Disconnected}

\begin{table}[H]
\caption{Optimal hyperparameters for training on the disconnected facts noise under minimum validation loss metric (left) and maximum validation accuracy metric (right).}
\label{Table.gh6}
\center
\includegraphics[width=\textwidth]{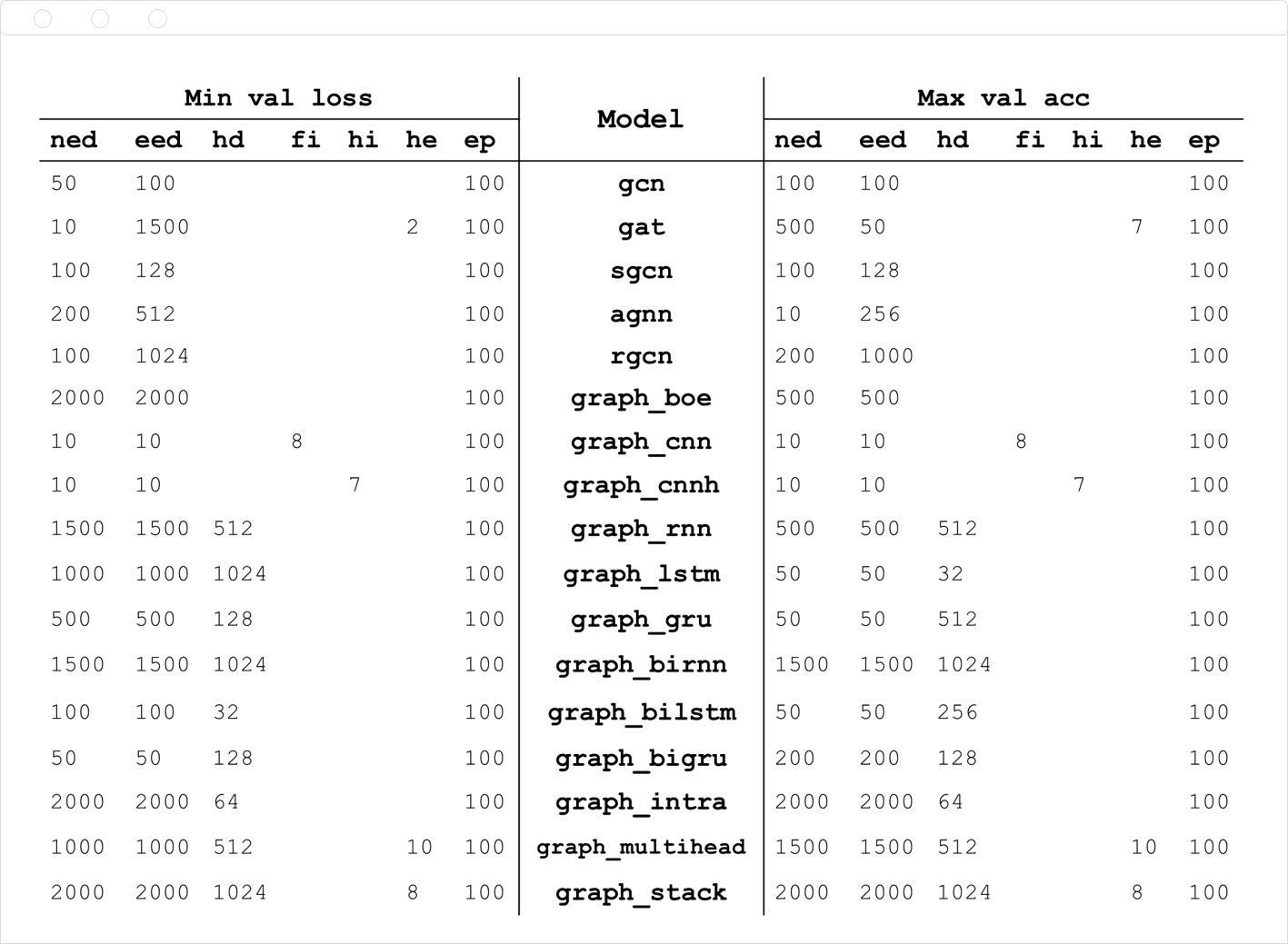}
\end{table}

\newpage
\begin{table}[H]
\caption{Training on the disconnected facts noise under minimum validation loss metric. The optimal values within the graph-based, the sequence-based and the text-based are in bold; the best performance among all models are in red.}
\label{Table.g61}
\center
\includegraphics[width=0.98\textwidth]{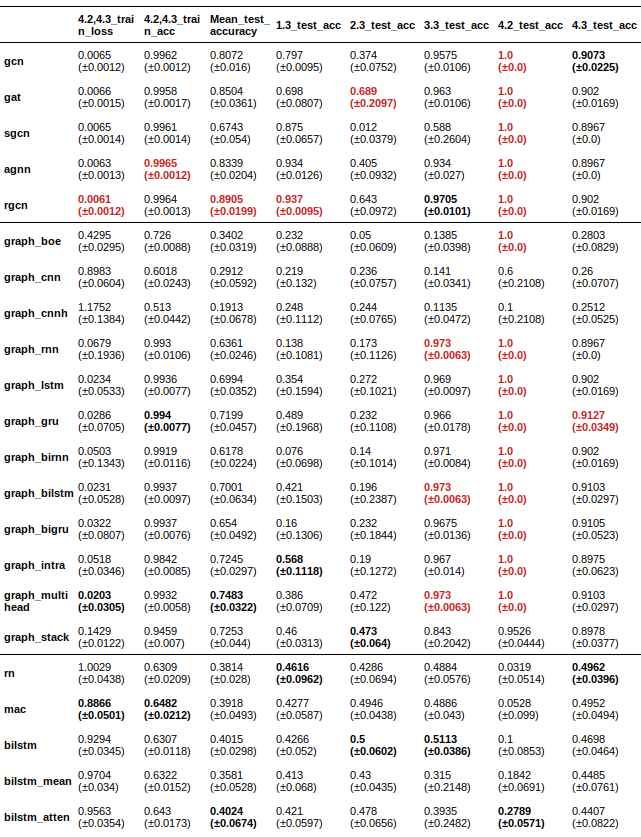} 
\end{table}

\newpage
\begin{table}[H]
\caption{Training on the disconnected facts noise under maximum validation accuracy metric. The optimal values within the graph-based, the sequence-based and the text-based are in bold; the best performance among all models are in red.}
\label{Table.g62}
\center
\includegraphics[width=0.98\textwidth]{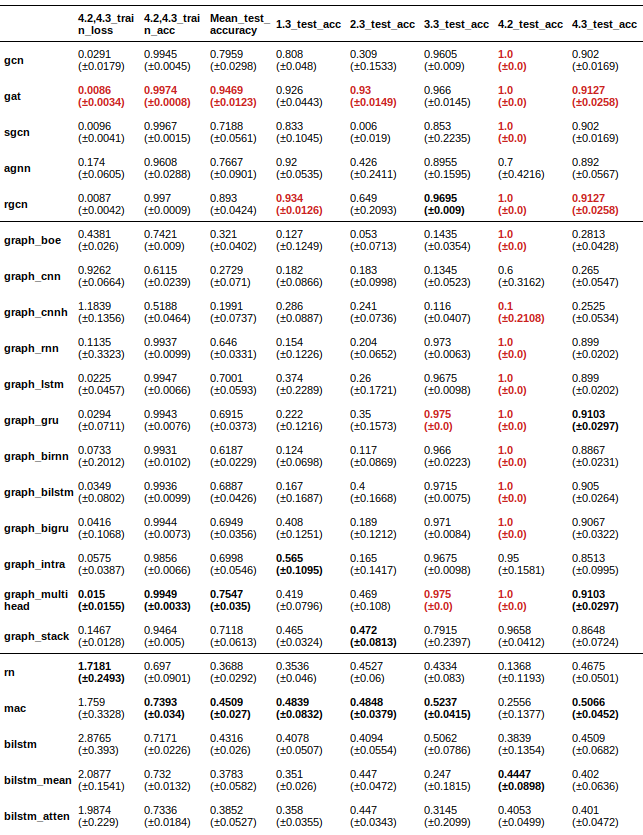} 
\end{table}

\end{document}